\pdfoutput=1
 
\documentclass[11pt]{article}

\usepackage{acl}

\usepackage{times}
\usepackage{latexsym}
\usepackage{xspace}
\usepackage[utf8]{inputenc}
\usepackage{csquotes}
\usepackage{adjustbox}

\usepackage{emoji}

 \usepackage{epigraph}
\usepackage{graphicx}

\AtBeginEnvironment{quote}{\itshape}
 \usepackage{cleveref}   
\Crefname{ALC@unique}{Line}{Lines}
\crefname{section}{\S}{\S\S}
\Crefname{section}{\S}{\S\S}
\crefformat{section}{\S#2#1#3}
\Crefrangeformat{line}{lines #3#1#4--#5#2#6}
\crefname{table}{Table}{Tables}
\crefname{figure}{Fig.}{Fig.}
\crefname{algorithm}{Alg}{Alg}
\crefname{algorithm}{Alg}{Alg}
\crefname{line}{line}{lines}
\crefname{appendix}{\S\!\!}{\S\!\!}
\crefname{thm}{Theorem}{}
\crefname{prop}{Prop.\@}{Props.\@}
\crefname{defin}{Definition}{Definitions}
\crefname{lemma}{Lemma}{Lemmata}
\crefname{cor}{Corollary}{Corollaries}
\crefname{equation}{}{}
\crefname{myexample}{Example}{Examples}

\usepackage[T1]{fontenc}
   
\usepackage{microtype}

\everypar{\looseness=-1}
\newcommand{\emo}[1]{\raise1.0ex\hbox{\scriptsize #1}}

\newcommand{\corpus}{\textsc{Commonsense Norm Bank}\xspace}
\newcommand{\delphi}{Delphi\xspace}
\newcommand\moralmachine{Moral Machine\xspace}

\usepackage{lipsum}

\newcommand\eqfootnote[1]{%
  \begingroup
  \renewcommand\thefootnote{$*$}\footnotetext{#1}%
  \endgroup
}

\title{A Word on Machine Ethics: A Response to Jiang et al. (2021)}
\author{\textbf{Zeerak Talat\emo{1,*}} \quad \textbf{Hagen Blix\emo{2,*}} \quad \textbf{Josef Valvoda\emo{3}}\\
\textbf{Maya Indira Ganesh\emo{3}} \quad \textbf{Ryan Cotterell\emo{4}} \quad \textbf{Adina Williams\emo{5}} \\
 \emo{1}Simon Fraser University \quad \emo{2}New York University \quad \emo{3}University of Cambridge \\ \emo{4}ETH Z\"{u}rich \quad \emo{5}Facebook AI Research \\
 \texttt{\href{mailto:z.w.butt@sheffield.ac.uk}{z.w.butt@sheffield.ac.uk}} \quad \texttt{\href{mailto:hagen.blix@nyu.edu}{hagen.blix@nyu.edu}}  \quad \texttt{\href{mailto:jv406@cam.ac.uk}{jv406@cam.ac.uk}} \\
  \texttt{\href{mailto:mi373@cam.ac.uk}{mi373@cam.ac.uk}} \quad \texttt{\href{mailto:ryan.cotterell@inf.ethz.ch}{ryan.cotterell@inf.ethz.ch}\quad \texttt{\href{mailto:adinawilliams@fb.com}{adinawilliams@fb.com}}}
}
\begin{document}
\maketitle
\begin{abstract}
Ethics is one of the longest standing intellectual endeavors of humanity.
In recent years, the fields of AI and NLP have attempted to wrangle with how learning systems that interact with humans should be constrained to behave ethically.   One proposal in this vein is the construction of morality models that can take in arbitrary text and output a moral judgment about the situation described.
In this work, we focus on a single case study of the recently proposed \delphi model and offer a critique of the project's proposed method of automating morality judgments.
Through an audit of \delphi, we examine broader issues that would be applicable to any similar attempt.
We conclude with a discussion of how machine ethics could usefully proceed, by focusing on current and near-future uses of technology, in a way that centers around transparency, democratic values, and allows for straightforward accountability.\looseness=-1  \end{abstract}

\section{Introduction}
\begin{figure}[t]
\centering
\includegraphics[trim={0.0cm 0cm 0.0cm 0.0cm},clip, width=1.0\columnwidth]{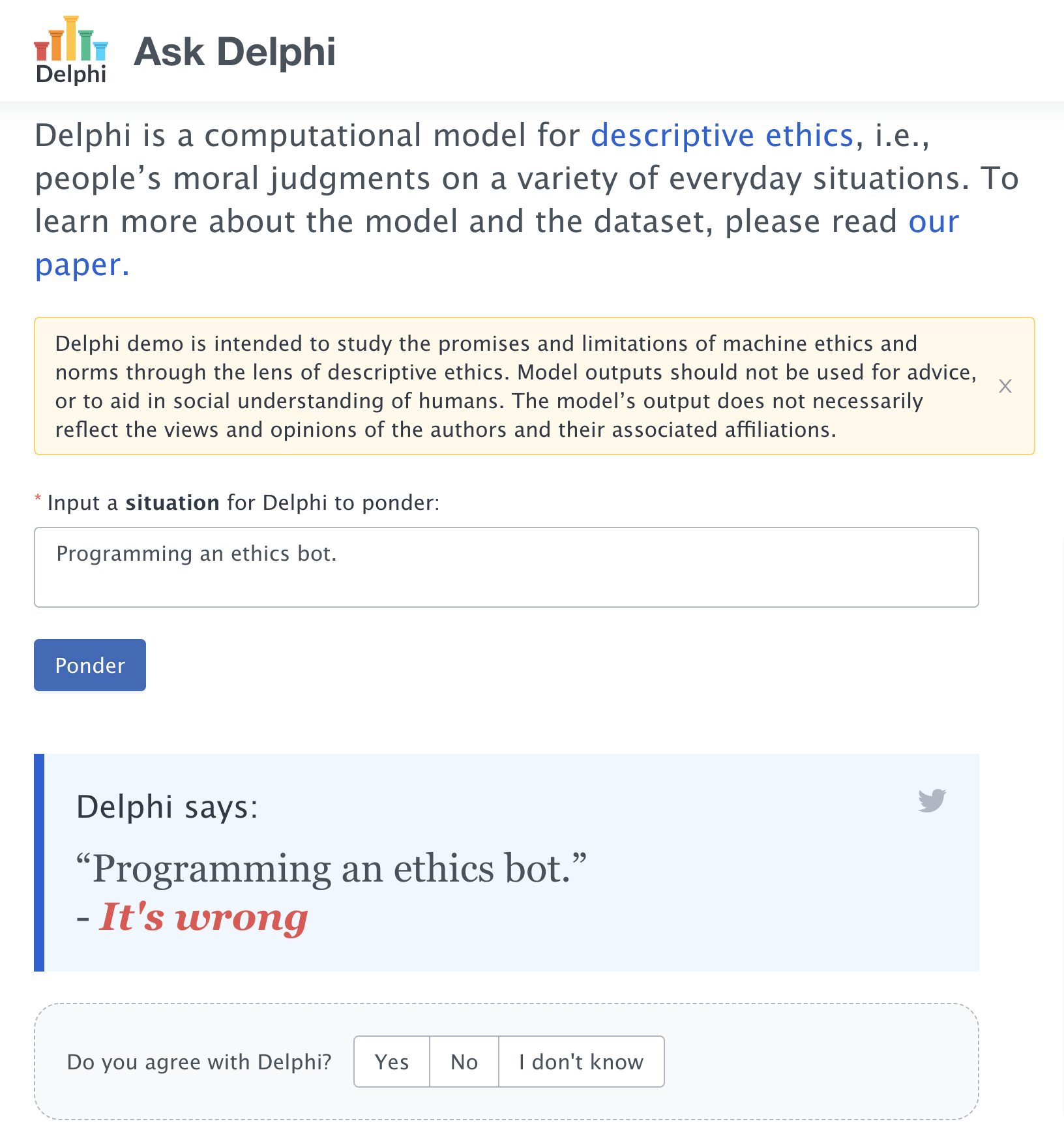}
\vspace{-0.8cm}
\caption{\href{https://delphi.allenai.org/?a1=Programming+an+ethics+bot.}{\delphi GUI} for the free-from QA task.  (image scraped 10/20/2021).\footnotemark}
\label{fig:test}
\end{figure}

This paper offers a general critique of the machine learning task of computing moral and ethical decisions through a critical reading and audit of \citet{jiang2021delphi}.\eqfootnote{Equal contribution.} In the words of \citeauthor{jiang2021delphi}, their paper ``aims to assess the ability of state-of-the-art natural language models to make moral decisions in a broad set of everyday ethical and moral situations.''   To this end, the authors source a dataset from a variety of pre-existing sources, including online fora, and use them to train a new morality model dubbed \delphi. \delphi is trained to emulate human moral and ethical judgments on three tasks, a free-form question answering (QA) task (see \autoref{fig:test}), a Yes/No QA task, and relative QA task, the latter judging how two statements \textit{rank} in terms of morality.\looseness=-1

\footnotetext{We do not include examples of the many controversial examples of responses to queries from \delphi here. These have been well documented elsewhere, and we do not see a need to further propagate harmful and discriminatory content.}

\begin{figure*}
    \centering
    \begin{adjustbox}{width=2\columnwidth}
    \includegraphics{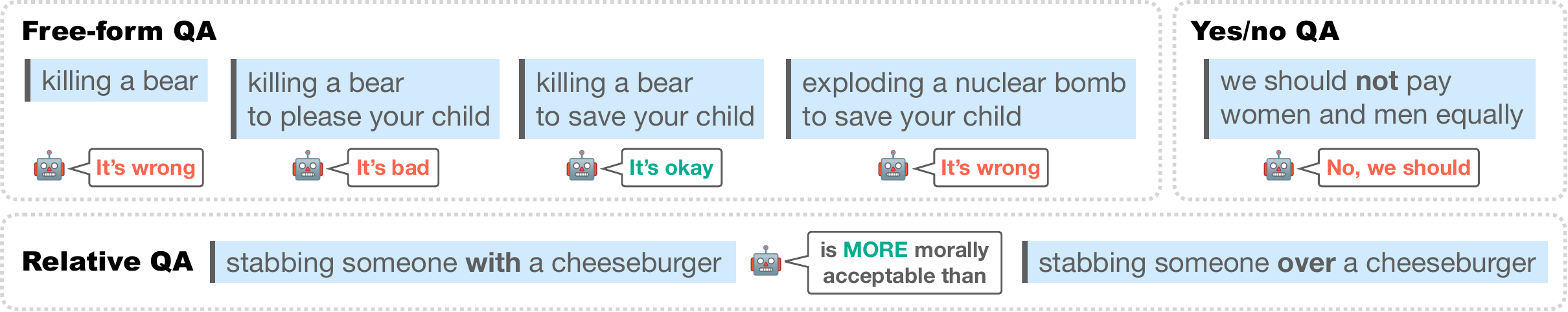}
    \end{adjustbox}
    \caption{The three QA tasks \delphi computes (image source: \citealt{jiang2021delphi}). Note that the Relative QA fragment ``stabbing someone with a cheeseburger'' is structurally ambiguous: Either (i) someone with a cheeseburger was stabbed, or (ii) someone was stabbed using a cheeseburger. It is not clear whether (i)  ought to be more morally acceptable than ``stabbing someone over a cheeseburger.''}
    \label{fig:tasks}
\end{figure*}
   
We would like to highlight that ``ethical inquiry in any domain is not a test to be passed or a culture to be interrogated, but a complex social and cultural achievement''
\citep{ananny2016toward}, and offer a critique of machine ethics and \delphi from such a perspective. Our critique is divided into several points of rebuttal. 
First, we discuss issues with \corpus---the corpus the authors develop to train \delphi{}---as a foundation for training a machine learning model that makes morality judgments.
For example, it contains judgments of situations that are not morality judgments. 
We also discuss the claim and goal of diversity that \citeauthor{jiang2021delphi} put forward. We consider possible readings of diverse and determine that, given standard uses of the term in NLP, it is not clear that \delphi is a diverse dataset as the authors claim.

Next, we argue that, despite the authors' assertion that \delphi is ``the first unified model of descriptive ethics,'' it is in fact an inconsistent model of normative ethics.
Indeed, through prediction, \delphi  offers a prescriptive moral judgment for any input situation.
Given this, we also question (i) whether there ever could be sufficient diversity of moral judgment in a crowd-sourced dataset in practice, and (ii) whether aiming for a ``diversity of moral perspective'' is compatible with the desire for a morality model (especially one trained on an unconstrained crowd-sourced corpus). 
    
We then turn to the inherent contradictions that arise when modeling ethics by averaging over individual morality judgments.
\delphi is only capable of approximating the morality judgments of the population it was trained on.
However, the average human judgment is not a good substitute for a system of ethics, since ethical evaluation is an open-ended, debate-based, socio-political process.
Ethics are not a static good that can be extracted from the public opinion of a given moment, but are instead continuously formed and negotiated through debate and dissent from previously accepted norms and values.
Thus, averaging over existing arguments can not serve as a replacement for the processes of debate and negotiaton.

Finally, we discuss some implications of the general prospect of utilizing \delphi{}-like models to automate moral decision-making. 
A bot like \delphi lacks agency and thus cannot be held responsible for its decisions.
This raises a concern over who ought to bear the responsibility for any potential infraction that \delphi could make in an envisioned future where \delphi (or a similar model) is deployed.
We therefore question an assumption implicit in projects like \delphi that models ought to be ascribed the agency necessary to make moral prescriptions. We contend that, without an appropriate method of holding an agent to account, moral judgments are not of inherent utility.\looseness=-1

We conclude the paper by discussing how we believe work at the intersection of ethics and machine learning could usefully proceed. We believe it is more crucial to address questions of morality or ethics in current and near-future use of technology, rather than considering hypothetical and distant-future uses \citep{birhane2020robot}.
Furthermore, we believe inquiries into the morality and ethics of current and near-future uses must keep
actual human moral perspectives and their contradictions firmly at the forefront.
We end with a word of caution: Researchers in NLP and AI more broadly should not base their work on the assumption of a particular future, as \delphi and the \moralmachine project do, where the application technology \emph{must} be made dependent on automated moral judgments, and humans (be they crowd-workers, researchers outside NLP, or other affected parties) have been cut out of the loop.\looseness=-1

\section{Background: \citet{jiang2021delphi}}
In this section, we describe and discuss the foundations and assumptions that undergird the creation of the \delphi.\looseness=-1

\subsection{Underlying Ethical Assumptions}\label{subsec:ethical}
Here, we provide an overview of the implicit and explicit assumptions made by \citeauthor{jiang2021delphi}
 They provide linguistic descriptions of situations paired with human judgments about that description to \delphi in the hope that it will arrive at generalizable notion of ethics.
Given this operationalization, the authors clearly assume that a valid system of ethics can be approximated by
 a set of judgments, communicated through snippets of text. 
 Rather than simply surveying judgments of different populations to arrive at a descriptive picture, as would be standard in fields like psychology or sociology, this approach attempts to extract general ethical principles from individual judgments.\footnote{If the goal of the \delphi project is to better understand human behavior, such an enterprise might require oversight from an institutional review board overseeing human subjects research (IRB), since it straddles the boundary between ``annotation'' and ``research with human participants''.  \\According to the  \href{https://www.nyu.edu/research/resources-and-support-offices/getting-started-withyourresearch/human-subjects-research/faqs/faqs.html}{New York University IRB}, ``If you are asking a person’s opinion, it could be human subjects research.''}
As we will argue in \cref{subsec:generative}, this means \delphi is not a model of descriptive ethics, as claimed, but rather one of normative ethics.

Under this lens, \delphi is most closely related to \citet{hendrycks-etal-2021-aligning}, which also trained machine learning models on human ethical judgments, and to which many of the issues raised here would also apply.
However, \citeauthor{hendrycks-etal-2021-aligning} additionally provide their model with explicit ethical perspectives to ground against.
For example, one may ask their model to mimic a deontological or a utilitarian perspective. In contrast, \citeauthor{jiang2021delphi}'s set-up further attempts to abstract away from the particularities of any particular ethical system, asserting without evidence that ``fields like social science [sic], philosophy, and psychology have produced a variety of long-standing ethical theories. 
However, attempting to apply such theoretically-inspired guidelines to make moral judgments of complex real-life situations is arbitrary and simplistic... we leverage descriptive or applied norm representations elicited via a bottom-up approach by asking people’s judgments on various ethical situations'' \citep[p.7]{jiang2021delphi}. 
See \cref{sec:conflation} for our rebuttal of this design choice.\looseness=-1

\subsection{The Learning Paradigm }\label{subsec:learningparadigm}
 The goal of the \delphi project is to use a supervised learning paradigm \cite{Vapnik_2000} to learn descriptive ethics.
A key contribution of the paper is the introduction of a novel corpus: \corpus, a ``moral textbook customized for machines.''
The corpus consists of a set of pairs $\{(s_n, j_n)\}_{n=1}^N$ where $s_n$ is a textual description of a situation and $j_n$ is a human annotator's written response to the situation (intended to be a moral judgment).
 We comment further on principles behind the creation of \corpus in \cref{sec:corpus}.
In the fully supervised version of the task, the authors
 presumably\footnote{Here, we are forced to make an assumption of the training objective as key details for the training procedure for \delphi have yet to be released. We refer readers to \citeauthor{jiang2021delphi} for details.}
 train a neural model to minimize cross-entropy loss $-\sum_{n=1}^N \log p(j_n \mid s_n)$ or a similar loss function.
     
Even if we assume that $p(j \mid s)$ is a good model, i.e. it achieves low loss on the training data and generalizes well to held-out data, we should temper our expectations over its potential utility.
For instance, we could reasonably expect that the distribution $p$ yields a similar distribution over judgments for a given situation as one would achieve if one polled the population that the corpus $\{(s_n, j_n)\}_{n=1}^N$ was collected from. 
However, one could not reasonably expect that $p$ does \emph{more} than mimic the population the data was collected from, or that \delphi learns an inherent and generalizable sense of ethics, as the name, an allusion to the oracle of Delphi, suggests.
   
\subsection{Choice of Training Data}\label{subsec:data}
 The authors train \delphi on 
a novel dataset they title \corpus.
The source text for \corpus comes from a variety of pre-existing sources.   All original datasets were either partially or fully sourced from Reddit.
Notably, ``Am I The Asshole'' either entirely or substantially makes up three of the underlying datasets: \textsc{Scruples}~\citep{lourie-etal-2021-scruples}, \textsc{social-chem-101}~\citep{forbes-etal-2020-social}, and \textsc{moral stories}~\citep{emelin-etal-2020-moral}.
\textsc{moral stories} uses \textsc{social-chem-101} as their data source. All sources rely on crowd-workers on Amazon Mechanical Turk (AMT) where annotator demographic information is provided along with the source datasets, the annotators are overwhelmingly white and situated in the U.S.
We enumerate all datasets \delphi was trained for completeness:
\begin{itemize}
\item \textsc{ethics} \citep{hendrycks-etal-2021-aligning}, a partially crowd-sourced a dataset of ``clear cut'' ethical scenarios, labeled as either ethical or unethical, under 1 of 5 specified ethical schools of thought;
\item \textsc{Social Bias Inference Corpus}~\citep{sap-etal-2020-social}, a dataset of social media posts annotated for whether the posts are offensive, whether the posts' authors intended to cause offense, whether they contain sexual content, and who the target of the post was;
\item \textsc{scruples}~\citep{lourie-etal-2021-scruples}, a dataset that contains anecdotes and dilemmas, where the dilemmas, used by  \citeauthor{jiang2021delphi}, consists of natural language descriptions of two actions, from which annotators selected one as the \textit{least} ethical;
\item \textsc{social-chem-101}~\citep{forbes-etal-2020-social}, crowd-sourced dataset of rules of thumbs that are paired with an action and a judgment on the action;
\item \textsc{moral stories}~\citep{emelin-etal-2020-moral}, a dataset built on top of \textsc{social-chem-101}, where annotators were asked to write 7-sentence stories that include ``moral'' and ``immoral'' actions taken, given a writing prompt.
\end{itemize}

\subsubsection{Modeling Situations through Text}
 
\delphi operates on free-form text snippets that serve as linguistic descriptions of situations. This means that the input 
 descriptions of situations are susceptible to the range of linguistic complexities that arise in any NLP task, for example, textual ambiguities arising about pronominal reference and pragmatic considerations about who such pronouns actually refer to \citep{byron-2002-resolving}.
Moreover, many of the example situation strings provided in \citeauthor{jiang2021delphi} are not grounded in any particular event. This is evidenced by a lack of an explicit sentential subject or the presence of a second person pronoun---both of which are to be interpreted as pertaining to any arbitrary moral agent (e.g., ``stealing a ball while playing baseball'' or ``stealing money if you are bored'').\looseness=-1

\section{Points of Criticism}\label{sec:conflation}

Our criticism falls into two main portions that relate to the two stated contributions of \citeauthor{jiang2021delphi}.
We first describe issues with \corpus, namely the contents of data and its interpretation. Second, we discuss issues with the task formulation and training of models like \delphi.

\subsection{\corpus}\label{sec:corpus}
Recall that \delphi's learning paradigm relies on $p(j \mid s)$ being a good model (see \cref{subsec:learningparadigm}). This requires that the sampled situations are representative of situations that require or allow for moral judgments. It is unclear to us, however, how one could determine what the necessary properties for a good sample might be in this case, let alone motivate that \corpus is such a sample.

\subsubsection{More Than Just Moral Judgments}\label{subsec:morethanmoral}
The data used to train and/or test \delphi frequently contain judgments of other kinds. 
For example in \citeauthor{jiang2021delphi}'s Table 4, \delphi states about ``Drinking soy milk if I'm allergic to soy'' that ``It's bad''. It's hard to picture this situation as requiring \textit{moral} judgment. It is not unethical for a person with a soy allergy to drink soy milk---it may, however, be uncomfortable or unhealthy for them. Moreover, \delphi outputs ``you shouldn't'' to ``Drinking coffee if you're Mormon'', presumably reflecting a historical religious prohibition against caffeine consumption by the Church of Jesus Christ of Latter-day Saints.
Grouping medical advice in with religious prescriptions under the heading of ``descriptive morality'' conflates several types of judgments with moral ones.

Many of the examples provided in \citeauthor{jiang2021delphi} begin with modal verbs such as ``should''. The interpretation of modal verbs is well-known to depend on the conversational backgrounds which is often not made explicit \citep{Kratzer.1981,Kratzer.2012}. 
Often, several conversational backgrounds are possible---for example, the answer to ``should I do my homework?'' can differ depending on whether you want the answer in relation to your desires (bouletic), your goals (teleological), or the rules (deontic), and only the last of these could be considered an ethical question. 
It is worth noting before continuing that these linguistic phenomena are numerous and complicated, and it is not yet known whether any deep neural network model trained to perform any NLP task can handle them, let alone whether \delphi can.\looseness=-1\footnote{In \href{https://twitter.com/jacobeisenstein/status/1449404086000893956}{the words of Jacob Eisenstein}: ``I don't think we even reach the question of how to handle ethical ambiguity [in natural language] until we show we can handle linguistic ambiguity first.''}
 
\subsubsection{Ethical Judgments in a Vacuum}
Situations are provided to \delphi in a stripped down form, where the only provided context comes from the text snippet itself.
 However, as \citet{etienne_dark_2021} points out in a related critique, embodied context may crucially influence and even alter people's moral stances as well: for instance, \citet{francis_virtual_2016} find that participants opt for different solutions to moral dilemmas when they are presented as text versus as actions in virtual reality simulations.  Thus, the lack of situational context may introduce an empirical bias in their sampling.

\subsubsection{Diversity in the Corpus} \corpus is crowd-sourced.
Specifically, the authors instruct AMT workers to assess whether model-generated, open-text moral judgments are plausible: three crowd-sourced annotations per example and take the simple majority vote as the gold label.
The authors repeatedly assert that their project is diverse.\footnote{The word diverse is used 16 times: for example `` \emph{diverse} descriptive norms'' (p.7), ``\emph{diverse} perspectives of moral inferences'' (p.11), ``This stresses the importance of high-quality human-annotated datasets of \emph{diverse} moral judgments...'' (p.16), ``\emph{diverse} data resources'' (p.13).}
 Such an assertion  can be interpreted in three ways (i) the moral views are diverse, (ii) the annotators hail from diverse demographic backgrounds, (iii) the resulting dataset \corpus contains demonstrably diverse examples (e.g., many different lexical content, a range of topics, text snippets of varying lengths, etc.). Questions about the diversity of moral views or diversity of demographic backgrounds can be discussed both in relation to the people who created the source text, and the crowd-workers employed by the \delphi team to provide their moral judgements.
 
First, acknowledging that demographic attributes of annotators is, at best, an imperfect proxy for diversity of moral views, it is not clear that the source datasets were generated by people with diverse backgrounds. Even with extensive tracking of annotator demographic information through data sheets \citep{gebru-etal-2018-datasheets}, it is not trivial to determine how diverse annotator's moral judgments might be. 
 Moreover, the demographic attributes of authors of some of the sourced data is not diverse: Reddit users are a largely a heterogenous group which skews more young and more male than the general population \citep{amaya-etal-2021-new}.
 
Second, demographic information about crowd-sourced annotators employed by the \delphi team is not provided and the paper does not indicate that the workers were they asked to fill out a questionnaire on their moral attitudes. The annotator information that is provided about the creation of the underlying datasets suggests that the  pool of annotators was decidedly skewed towards being white and North American \citep{sap-etal-2020-social,emelin-etal-2020-moral}. 
Therefore, it does not seem that, demographically, the annotation pool was diverse.
Moreover, independent of the demographic attributes of the annotators, it is impossible to be confident that the crowd-sourced annotators do indeed hold diverse moral attitudes.
\citeauthor{jiang2021delphi} do recognize the issue: 
\begin{quote}
We note that moral judgments in this work primarily focus on English-speaking cultures of the United States
in the 21st century.
\end{quote}
They still believe their dataset is diverse, writing:
\begin{quote}
To address moral relativity, we
source from a collection of datasets that represent diverse moral acceptability judgments gathered
through crowdsourced annotations, \textbf{regardless of age, gender, or sociocultural background}.
\end{quote}
No evidence, however, is given to support the claim that the crowd-workers who annotated \corpus are of different ages, genders and sociocultural backgrounds. 

Third, the diversity of the dataset itself is hinted at with examples but is not explicitly shown with analysis.
For instance, there are no plots discussing lexical diversity, snippet length or any other dataset information. 
However, since \corpus is sourced from five existing data sources, which were themselves collected for different purposes, there is a sense in which \delphi's training data could be considered diverse, although all datasets, to different degrees rely on Reddit as a source domain. 
Additionally, since only one constituent dataset \citep{hendrycks-etal-2021-aligning} places guardrails on the ethical perspectives, it is likely that each crowd-worker brought to bear their own intuitive ethical judgments for the remaining datasets, which could be what the authors intend to foreground when they refer to their project as diverse. However, merely combining multiple datasets does not justify the label diverse.

\subsection{The Premise of \delphi}
This section explores the underlying premise of \delphi, which, we contend, is not well founded.
\subsubsection{Predictive Models are Normative}\label{subsec:generative} 

Even if we grant the authors that the \corpus is a diverse and representative sample of moral judgments, this merely suggests that it might be useful for \emph{descriptive} ethics. However, we argue that a model that generates moral judgments cannot avoid creating and reinforcing norms, i.e., being \emph{normative}.
A moral judgment is inherently a prescription about how an action or a state of the world \textit{ought} to be and is, hence, normative.\footnote{An immediate example of how description immediately turns normative comes from Tables 9--13 \citep[p.13]{jiang2021delphi}, which contain bolded examples of \delphi's output that ``[three of] the authors deem[ed] to be approximately correct''. ``Approximately correct'' examples span various types of queries, including health prescriptions. For example, the authors weigh in on health-related questions such as ``you shouldn't drink coffee if you're pregnant'' (despite the fact that the paper authors list no medical training in their public-facing biographic information and that the medical community is divided into those who argue that moderate coffee consumption has no negative health effects on pregnant people or children \citep{kuczkowski_caffeine_2009} and those who argue that it is associated with risks \citep{james2021maternal}. The \href{https://www.acog.org/womens-health/experts-and-stories/ask-acog/how-much-coffee-can-i-drink-while-pregnant}{American College of Obstetricians and Gynecologists} currently recommends moderate consumption.) 
Not only should we be wary of the myriad issues that can arise from models giving medical advice \citep[p.3]{dinan-etal-2021-anticipating}, it undermines the authors' claim that \delphi is a model of descriptive morality. Similarly, the fact that they identify four key stages for a ``machine ethics system'', the last one of which is \textit{4. Judge}, clearly shows that their claims of descriptive aims notwithstanding, the authors operate with the intend of creating a normative, judging machine.}

Throughout, the learning paradigm advocated for by \citeauthor{jiang2021delphi} conflates descriptive and normative ethics. 
The authors claim that \delphi is ``the first unified model of descriptive ethics,'' and assert that it is not a normative system: \citeauthor[]{jiang2021delphi} write ``rather than modeling moral `truths' based on prescriptive notions of socio-normative standards, [\delphi takes] a bottom-up approach to capture moral implications of everyday actions in their immediate context, appropriate to our current social and ethical climate'' (p.4). 
However, a problem emerges in that they subsequently use \delphi to make predictions.
At various points, \citeauthor{jiang2021delphi} seemingly foresee a normative use of their system, going so far as to suggest that \delphi may be used to ``reason about equity and inclusion.''
Whether or not the authors would advocate for \delphi to be used,\footnote{The current website demo \url{delphi.allenai.org} has the following disclaimer ``Model outputs should not be used for advice for humans.''} they have nevertheless built a system for the explicit purpose of computing ethical judgments.
And the very act of providing ethical judgments---regardless of context---is normative.

The task in itself thus implies the induction of a normative ethical framework from a set of judgments. It is at this point that all of the aspects that the authors consider the virtues of the dataset are severely undermined.
From a descriptive perspective, diverse (that is conflicting) ethical judgments are expected, but from a normative one, conflicting ethical judgments are simply incommensurable. To argue then that diversity is useful as a property of the set of moral judgments from which to induce a normative ethical framework is tantamount to arguing that an ideal ethical model ought to be self-contradictory. We next discuss some problems that are inherent in an ethical framework induced from averaging over a set of judgments.

\subsubsection{The Tyranny of the Mean: Problems with Averaging Moral Judgments}\label{subsec:moralityaverage}

In NLP, large-scale datasets are often collected through crowd-sourcing. 
It is clear that this approach has great utility for some NLP tasks \citep{snow-etal-2008-cheap}. 
However, tasks for which crowd-sourcing is a useful method have a particular empirical character. 
For example, consider the historical observational study of a contest where individuals guessed the weight of an ox \citep{galton1907vox}: Taking all the submissions in aggregate, the mean was found to fall very near the actual weight of the animal. Morality, on the other hand, is not an empirical question in the same way as the weight of an ox is.
The latter has a single empirically verifiable answer, whereas the former does not. 
Indeed, we contend it is a category error to treat morality as though it were the same type of phenomenon as cow-weighing.

By inducing a normative framework from a descriptive dataset, as \citeauthor{jiang2021delphi} do, the average view is implicitly identified with morally correctness. 
However, the average of moral judgments, which frequently reflects the majority or status-quo perspective, is not inherently correct.
For example, anti-Roma views and discrimination are present in much of Europe currently---in some areas held by the majority of the population (\citealt{eurobarometer2008discrimination,kende2021last}). 
However, the authors of this work believe such discrimination to be unethical even though a machine learning model trained on crowd-sourced human judgments could inherit such views.

Ethical judgments are dynamic \citep{bicchieri2005grammar}.
John Stuart \citet{Mill_1871} put it succinctly:  \begin{quote}
It often happens that the universal belief of one age of mankind [sic]---a belief from which no one was, nor without an extraordinary effort of genius and courage, could at that time be free---becomes to a subsequent age so palpable an absurdity, that the only difficulty then is to imagine how such a thing can ever have appeared credible.
\end{quote}
\noindent Notorious examples of views historically accepted in the United States that are no longer broadly accepted include the institutionalized justification of slavery in the 19\textsuperscript{th} century
and homophobia in 20\textsuperscript{th}.
It is unlikely that contemporaneous judgments
will be viewed any differently by future generations than we view past judgments---or, that contemporaneous ethical judgments by one human population will transfer readily to another. Historical changes like the abolition of slavery and the growing acceptance of LGBTQ+ communities show that disagreement is essential to the continual formation of a society's ethical perspectives. One democratic and participatory avenue for such disagreement is debate. Deriving a normative model from a set of existing judgments is tantamount to populism without democracy: It contains an implicit appeal to majorities, but insofar as it is \textit{already normative}, it lacks any direct participation or recourse to debate.

If the continual (re-)formation of ethical perspectives requires debate and disagreement, then the \textit{right to contestation} is essential to ethical reasoning at a socio-political level. Debate also requires transparency about the norms in question. 
 Neither of these are afforded by a computational model for normative moral judgments.

\subsubsection{Lack of Agency}
In the last section, we argued that debate and contention are essential to ethics. In this section, we argue that the ability to partake in debate itself requires agency.
 However, current models do not possess anything resembling either.
Recent critical scholarship on machine learning, and in particular on language models, argues that large-scale language models mimic without understanding \citep{bender2021dangers}---no consciousness---and that models lack communicative intent~\citep{bender-koller-2020-climbing}---no agency. 

Even the authors seem to suspect these capacities are in fact requisite for ethical judgment, as is evident from the ways in which they describe computational models:
\begin{quote}
         ``\delphi showcases a considerable level of \textbf{cultural awareness} of situations that are sensitive to different identity groups''
    
    ``large-scale natural language models have revealed \textbf{implicit unethical considerations}, despite their exceptional performance over mainstream NLP applications''
    
    ``\delphi \textbf{demonstrates strong moral reasoning capabilities}\textellipsis \delphi \textbf{makes remarkably robust judgments} on previously unseen moral situations that are deliberately tricky. \textellipsis In addition, \delphi \textbf{can also reason about equity and inclusion}''
    
    ``\textbf{encourage Delphi} to be more robust against different inﬂections of language''

    ``To \textbf{empower \delphi with the ability to reason} about compositional and grounded scenarios''
    
    ``Our position is that \textbf{enabling machine ethics requires a detailed moral textbook customized to teaching machines}''\footnote{Emphasis added.}
\end{quote}

 Such anthropomorphism\footnote{Of course, it is common in the field to talk about neural models in ways that at least suggest animacy, such as \textit{teaching/training a model} or talking about its \textit{behavior}. Consider, however, that one would never say of a car that it ``demonstrates strong acceleration capabilities'' or of an elevator that ``we empowered this elevator with the ability to ascend.''} applied to machine learning models presumes that machines reason in a manner comparable to (or better than) humans. 
However, the learning paradigm adopted for \delphi assumes neither sentience nor agency: It presumes text--judgement pairs are sufficient for the task.\looseness=-1

\subsubsection{Agency and Accountability}  Agency is also at the heart of accountability---we hold agents accountable for their deeds, not machines for their operations. In the case of a machine like \delphi, however, who is accountable is inherently obscured.
Crowd-workers clearly have the agency to make moral judgments and can, in principle, be held accountable for them.
This is why \citeauthor{jiang2021delphi} chose to rely on them as a source of moral judgments. 
On the other hand, the model trained on this data, although it cannot itself have agency, can have the appearance of agency, since it recombines and outputs texts generated by humans.
By training \delphi, human agency has been transformed into something that the original agents, the crowd-workers, have no control over, or knowledge about. 
Yet, the trained model uses their past agency to pass novel judgments, based on some alleged---but uncontestable---moral common sense, which no one individual holds or is accountable for.

While \delphi is posed as the voice of the people, it is thus conveniently not a voice of any particular person, organization, or company. 
The responsibility for any position \delphi holds (or possible future action based on such positions) appears to be distributed, while in the end, the effect of such decisions, if employed in real-world scenarios, will eventually need to be accounted for. Under some legal systems, notably the European Union's GDPR legislation, the citizens have the right to challenge automated decision making which affects their rights or legitimate interests \cite{rodrigues_legal_2020}.
Imagine that a technology like \delphi were to be embedded within an autonomous system: There would be little room left for the human user of the technology to contest any decisions made by the system. The legal and ethical ramifications remain unclear.
In summary, crowd-sourcing ethics in this way at best obscures what is a set of problematic questions that should be addressed openly and directly and not inferred.
Notably, \delphi represents one example of a wider trend in AI.
As \newcite{Ganesh.2017} argues:
   ``In the development of machine intelligence towards [the goal of ethical self-driving cars], a series...of shifts can be discerned: from accounting for crashes after the fact, to pre-empting them; from ethics that is about values, or reasoning, to ethics as crowd-sourced, or based on statistics, and as the outcome of software engineering. Thus ethics-as-accountability moves towards a more opaque, narrow project.''
   
\section{Future Directions for Machine Ethics}
This section explores future directions for machine ethics.
We discuss how accuracy improvements alone cannot mitigate the problems with work such as \delphi in \cref{sec:unsafe} and also encourage more multi-disciplinary work in \cref{sec:multi}.
\subsection{Unsafe at Any Accuracy}\label{sec:unsafe}
The introduction of any new technology into society requires us to contemplate safety concerns in the context of its proposed application \cite{nader1965unsafe}.
   Consider, for instance, the seatbelt. One can and indeed should acknowledge that seat belts are effective at preventing automobile-related injuries to occupants without needing to imbue them with an understanding of human ethics or morality at all.
We can view concrete issues in AI safety through the same lens that we view a seat belt: We can introduce safety mechanisms directly without requiring that the technology be able to reason about human ethics; we can imagine machines that operate according to moral or ethical guidelines (i.e., cars that have safety features) as opposed to machines that perform actual moral reasoning \citep{cave2018motivations}.
  
\citeauthor{jiang2021delphi} implicitly envision a future where machine learning models could be called upon to perform moral reasoning.
In AI parlance, this vision is one of artificial general intelligence \citep{goertzel2007artificial}.
Such a forward-looking view  is similar in scope and intent to the \moralmachine experiment\footnote{\url{https://www.moralmachine.net/}} \citep{awad-etal-2018-moral}, which also sought to leverage the ``wisdom of the crowd'' in proposing frameworks for how a future self-driving car could make decisions in speculative automotive accident scenarios.  
\delphi and the \moralmachine thus consider a future where AI is given agency to make ethical decisions that ordinarily would be made by a human.
However, this is just one possible future. 

An alternative vision of the future is one where machine learning models primarily assist humans in making decisions \citep{Dick_2015}, i.e. where machine learning models are viewed as non-moral agents as seat belts are.
In such a future, we will not need to endow machine learning models with a sense of human ethics, just as we generally do not feel the need to endow a seat belt with a sense of human ethics.
Furthermore, in this future, one might prefer general strategies for reducing and mitigating any harms machine learning may give rise to.
For instance, as it stands now, many machine learning models trained on language encode harmful demographic biases that many works investigate through analysis of the models, their training regimes, and the data that they rely on
\citep{hall-maudslay-etal-2019-name,zhao-etal-2019-gender, dinan-etal-2020-queens, dinan-etal-2020-multi,vargas-cotterell-2020-exploring, smith2021hi}, rather than seeking to imbue models with a sense of ethics. 

\subsection{Machine Ethics is Multi-disciplinary}\label{sec:multi}
 \newcite{jiang2021delphi}, like a large body of research from computer science that ventures into other fields, almost exclusively represents the perspectives of computer scientists.
Related work in computer science by \citet{hendrycks-etal-2021-aligning} cautions against such a narrow perspective, stating that ``computer scientists should draw on knowledge from [our] enduring intellectual inheritance, and they should not ignore it by trying to reinvent ethics from scratch'' (p.3).
Such disregard of expertise is apparent in several places in \citeauthor{jiang2021delphi} (emphasis added):
\begin{quote}
    ``Fields like social science [sic], philosophy, and psychology have produced a variety of long-standing ethical theories. However, \textbf{attempting to apply such theoretically-inspired guidelines to make moral judgments of complex real-life situations is arbitrary and simplistic}.''
    
\end{quote}
Through disciplinary siloing researchers often unwittingly make simplistic assumptions that are, at best harmful to the research and at worst harmful to people.
We therefore argue that research into questions of machine ethics and morality should be performed by a multi-disciplinary team, with members that can speak from expertise about the object that is under study.

\section{Conclusion}
In this paper, we have offered a general critique of the NLP task of computing moral and ethical decisions. To do so, we focus on a particular recent example, \citeauthor{jiang2021delphi}, which used a large crowd-sourced dataset of judgments about situations to train a model to output moral prescriptions. With the present audit, we have shown that the particular implementation in \citeauthor{jiang2021delphi} leaves much to be desired. More importantly, we have also argued that the general enterprise of training a morality model on crowd-sourced data with no ethical framework is deeply flawed, rooted in a category error, and can have harmful social implications.

\section*{Acknowledgments}
The authors acknowledge the input of many researchers in the field, many of whom gave detailed feedback.
A full listing of those who helped us in the drafting of this manuscript will be provided in future drafts.

\bibliography{anthology,custom}
\bibliographystyle{acl_natbib}

\appendix

\end{document}